\documentclass[final]{cvpr}
\usepackage{multirow}
\usepackage{times}
\usepackage{epsfig}
\usepackage{graphicx}
\usepackage{amsmath}
\usepackage{amssymb}
\usepackage{booktabs}
\newcommand*{\affaddr}[1]{#1} 
\newcommand*{\affmark}[1][*]{\textsuperscript{#1}}
\newcommand*{\email}[1]{\texttt{#1}}

\usepackage[pagebackref=true,breaklinks=true,colorlinks,bookmarks=false]{hyperref}

\begin{document}

\title{\ Learning to Filter: Siamese Relation Network for Robust Tracking }

\author{
Siyuan Cheng\affmark[1,2], Bineng Zhong\affmark[1]\thanks{Corresponding author.}, Guorong Li\affmark[3], Xin Liu\affmark[4], Zhenjun Tang\affmark[1], Xianxian Li\affmark[1*], Jing Wang\affmark[2]\\
\affaddr{\affmark[1]Guangxi Key Lab of Multi-Source Information Mining \& Security, \\ Guangxi Normal University, Guilin 541004, China}\\
\affaddr{\affmark[2]Department of Computer Science and Technology, Huaqiao University, China}\\
\affaddr{\affmark[3]School of Computer Science and Technology, University of Chinese Academy of Sciences, China},\\
\affaddr{\affmark[4]Seetatech Technology, Beijing, China}\\
\email{siyuancheng@stu.hqu.edu.cn},
\email{bnzhong@gxnu.edu.cn},
\email{liguorong@ucas.ac.cn}\\
\email{xin.liu@seetatech.com},
\email{rrji@xmu.edu.cn},
\email{tangzj230@163.com},\\
\email{lixx@gxnu.edu.cn},
\email{wroaring@hqu.edu.cn}
}
\maketitle

\pagestyle{empty}
\thispagestyle{empty}

\begin{abstract}   
    Despite the great success of Siamese-based trackers, their performance under complicated scenarios is still not satisfying, especially when there are distractors. To this end, we propose a novel Siamese relation network, which introduces two efficient modules, i.e.\ Relation Detector (RD) and Refinement Module (RM). RD performs in a meta-learning way to obtain a learning ability to filter the distractors from the background while RM aims to effectively integrate the proposed RD into the Siamese framework to generate accurate tracking result.
    Moreover, to further improve the discriminability and robustness of the tracker, we introduce a contrastive training strategy that attempts not only to learn matching the same target but also to learn how to distinguish the different objects. Therefore, our tracker can achieve accurate tracking results when facing background clutters, fast motion, and occlusion. Experimental results on five popular benchmarks, including VOT2018, VOT2019, OTB100, LaSOT, and UAV123, show that the proposed method is effective and can achieve state-of-the-art results. The code will be available at \url{https://github.com/hqucv/siamrn}

\end{abstract} 

\section{Introduction}   

Visual object tracking, which is the fundamental task in computer vision, has received much attention over the last decades~\cite{kcf, weakly, hierarchical, deepalignment}. It aims to capture the position of an arbitrary target accurately by given only its initial state~\cite{1survey}. With the increasing demand for the practical application such as autonomous driving~\cite{autodrive}, robotics, surveillance~\cite{surveillance} and human-computer interaction~\cite{human_computer}, current trackers require not only accuracy but also speed and robustness to overcome the existing challenges, such as occlusions, fast motions, appearance deformations, illumination change, and background clutters~\cite{challenge}, etc.
\begin{figure}
\begin{center}
\setlength{\fboxrule}{0pt}
\setlength{\fboxsep}{0cm}
\fbox{\includegraphics[width=1\linewidth]{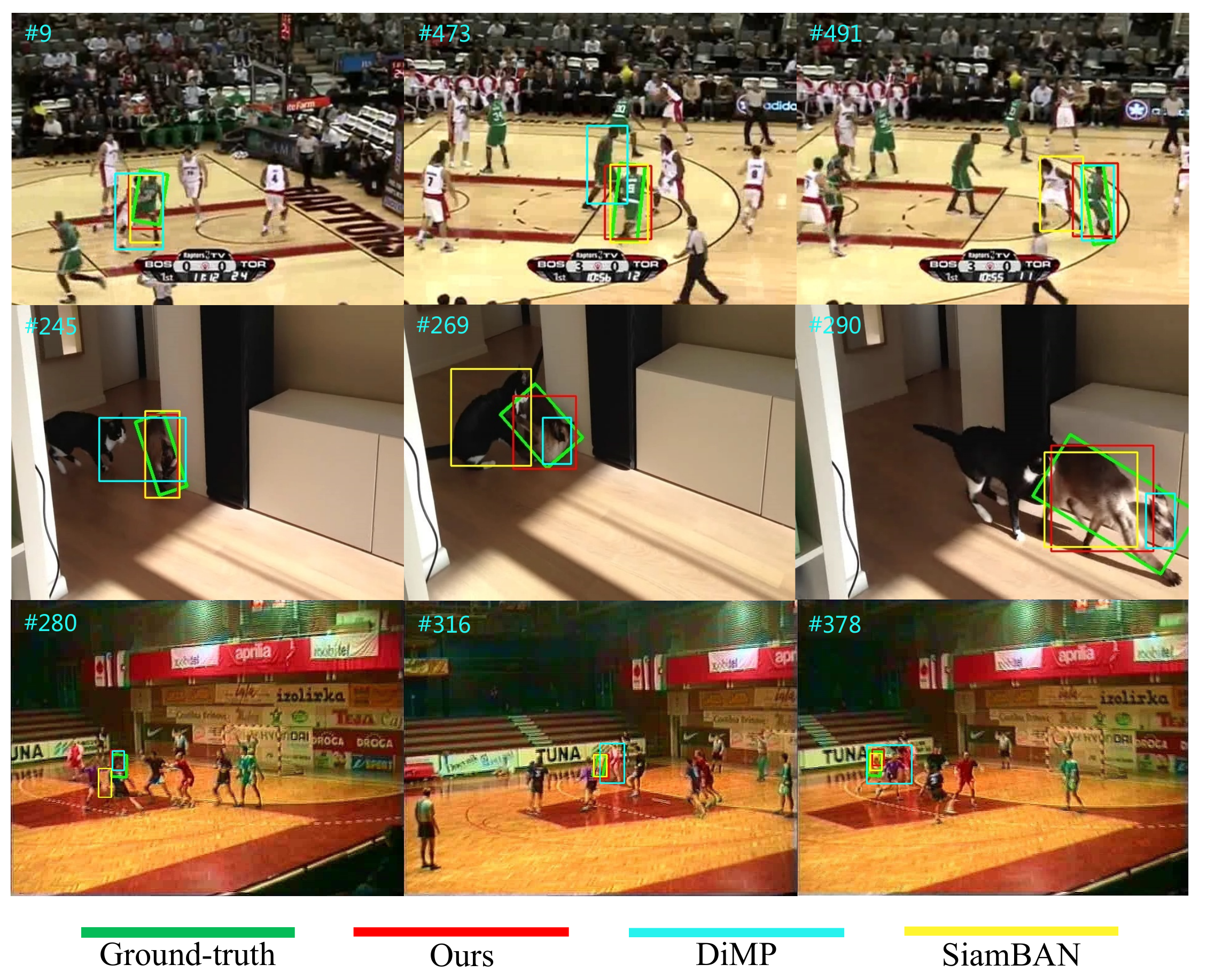}}
\end{center}
  \caption{ Tracking Result of our Siamese relation network with two state-of-the-art trackers in VOT2018~\cite{vot2018}. Benefiting from our Relation Detector and Refinement Module, when facing similar distractors, appearance change and complex background, our tracker is more robust to these challenges and gets more accurate results. Figure \ref{fig:heatmap} further shows the ability to filtering distractors from the target region.}
\label{fig:compare_box}
\end{figure}

    Owing to the development of CNN and the various network architectures, we can unveil the powerful deep feature for computer vision tasks ~\cite{cnn}. In recent years, the Siamese network based trackers~\cite{siamban, siamfc, siamrpn, dasiamrpn, siamrpn++} have drawn great attention due to their balanced speed and performance. 
    However, there still exists a gap of robustness between the Siamese trackers and the discriminative trackers that are equipped with an update mechanism~\cite{dimp, eco, updt, ladcf, vot2018}, which can be manifested in the occasion of tracking failure when encountering similar distractors or background clutters. 
    This is because the training setting of the Siamese trackers is only to match the same target in abundance of image pairs and ignore to distinguish their difference. Therefore, the discriminability for Siamese trackers is deficient when addressing complicated situations.
    Meanwhile, the target is incompetent to be highlighted due to the Siamese-style cropping strategy, which may introduce the distractive context.
    Moreover, the classification and regression branches, which are two significant pillars in Siamese-based trackers, are usually optimized independently ~\cite{siamban,siamrpn++,siamrpn}, increasing the probability of mismatch between them during tracking. To be specific, the box corresponding to the position with the highest classification confidence is not the most accurate one for the tracked target.



To address the above issues, we propose a novel Siamese relation network, which introduces two efficient modules named as Relation Detector (RD) and Refinement Moudle (RM). RD aims to learn the ability to filter the distractors from background via meta-learning. Specifically, we define two categories labeled as target and non-target. Given the training pairs assigned with the label, we perform a two-way one-shot learning, which intends to measure the relationship between them by the nonlinear learnable comparators. Therefore the distractors are filtered out due to low-relevant relationships. The RM is used to integrate the information obtained by RD and the classification branch to refine the tracking results, alleviating the mismatch between classification and regression branches.
Furthermore, we equip RD with a contrastive training strategy that attempts not only to learn matching the same target but also pay attention to distinguish the different objects by forming a training triplet and the different combinations of samples, which effectively boosts the discriminant ability of the proposed tracker.
Benefiting from the above modules, our tracker possesses remarkable discriminability and robustness when facing background clutters, fast motion, and occlusion, as shown in Figure~\ref{fig:compare_box}. Experiments (details in Section~\ref{section:implement}) show that our tracker achieves state-of-the-art performance on five popular tracking benchmarks, which confirms the effectiveness of the proposed method.

Our main contributions can be summarized as follows.

\begin{itemize}

\item We introduce a novel Relation Detector (RD) that is trained to obtain the ability to filter the distractors from background via few-shot learning based contrastive training strategy. Benefit from RD, during the tracking procedure,  our tracker can distinguish the target in the cluttering background once given the initial state of the target without further fine-tuning. 

\item To integrate the information obtained by RD and the classification branch to refine the tracking results, we design a Refinement Module, which can jointly operate the classification and regression to localize the target, reducing the mismatch between those two branches. 

\item Our method achieves state-of-the-art results on five popular benchmarks, including VOT2018, VOT2019, OTB100, LaSOT, and UAV123, which confirms the effectiveness of our proposed method.

\end{itemize}

   \begin{figure*}
\begin{center}
\setlength{\fboxrule}{0pt}
\setlength{\fboxsep}{0cm}
\fbox{\includegraphics[width=.8\linewidth]{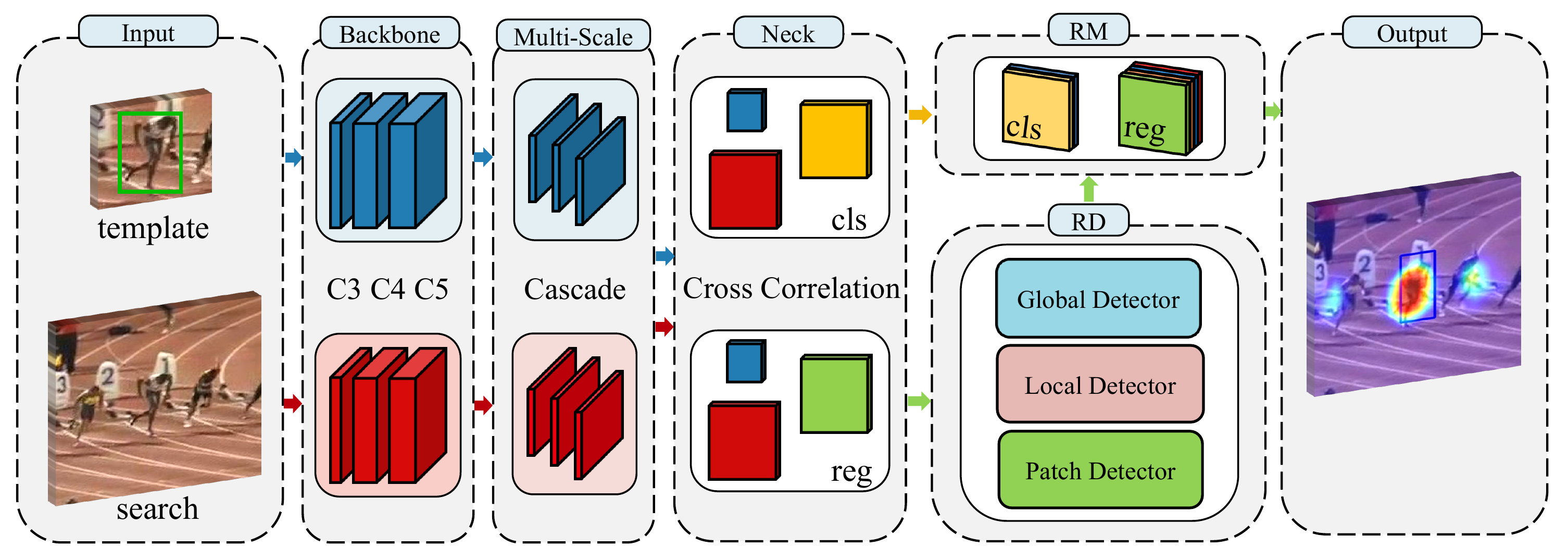}}
\end{center}
  \caption{The tracking pipeline of the proposed Siamese Relation Network. The proposed Relation detector (RD) and joint Refinement Module (RM) are presented. During tracking, we feed the features of proposals that are generated from the regression branch by precise ROI pooling~\cite{atom} into RD to measure the relationships with target-specific feature. Then we convert the output of the RD to a matching score and utilize it in the Refinement Module (RM) to jointly operate the regression and classification branches to predict the target location.}
\label{fig:pipeline}
\end{figure*}


\section{Related Work}


\subsection{Siamese network based Trackers} 
Recently, Siamese network based trackers consider tracking as a metric learning problem and have drawn great attention for their good trade-off between speed and accuracy~\cite{dsiam, cfnet,dasiamrpn, siamrpn, siamrpn++, siamban}. Bertinetto \etal~ first introduced SiamFC ~\cite{siamfc} for visual tracking, which aims to learn the similarity metric between the target template and search region and operate a comparison formed as the cross-correlation to localize the target. 
With the emergence and success of region proposal network (RPN)~\cite{faster_rcnn}, Li \etal~~\cite{siamrpn} applied it into the Siamese networks framework, referred to as SiamRPN, which solved the scale variation of the target. In order to unveil the powerful deep feature extracted by the deep network, SiamRPN++~\cite{siamrpn++} and SiamDW ~\cite{siamdw} tackled the challenge brought by introducing the deep network into the Siamese framework and made the performance highly improved.
Chen \etal~ \cite{siamban} borrowed the idea from FCOS~\cite{fcos} into tracking and designed a simple yet effective anchor-free tracker, which can get rid of the intricate parameters of anchor setting.
However, these Siamese-based trackers usually localize the target by using the classification and regression branch and optimize them independently, which can arouse the mismatch between them in tracking procedure~\cite{mal}. Alternatively, we design a Refinement Module that can jointly operate the optimization of these two branches by collaborating with RD output, which balances these two branches to obtain more confident and accurate tracking results.

\subsection{Backgroud Distractors in Visual Tracking}
The Siamese network based trackers are susceptible to be affected by background distractors, which severely hampers the robustness of tracker. To find out its causes is the less capable discrimination ability. DaSiamRPN~\cite{dasiamrpn} finds the inappropriate distribution of semantic and non-semantic negative samples hamper the discriminability of the tracker, so they replace one of the image in training image pairs by negative semantic sample randomly. Recently, some approaches aim to fuse online updating strategy into Siamese framework to pursue more discrimination performance. CFNet~\cite{cfnet} interpreted the correlation filter as a differentiable layer to update the model online, and DSiam~\cite{dsiam} used a fast transformation module to perform online learning.
UpdateNet~\cite{updatenet} tried to estimate the target template during tracking to overcome the target appearance variation.
However, online update strategy needs heavy computation and the accumulated tracking error will eventually lead to failure. Alternatively, we proposed a Relation Detector (RD), which learns to filter the distractors from background, and further develop a contrastive training strategy that not only to learn matching the same target but also to learn how to distinguish the different objects. Our model is trained offline and can advance the discriminability and robustness of the tracker stably when facing complicated scenarios.


\subsection{Few-Shot Learning and Meta Learning}   
Few-shot learning aims to recognize novel visual categories from very few labeled examples~\cite{few-shot}. 
There is usually only one or very few feasible data during training. Therefore how to overcome this situation to ensure the generalization of the model becomes a challenging task~\cite{few-shot1}. The mainstream few-shot learning approaches are optimized by recurrent neural network (RNN)~\cite{DBLP:rnn1}, further fine-tuned on the target problem~\cite{maml} or trained to learn an effective metric~\cite{metric1,learning_compare}. A popular trend is to design a general strategy that can guide supervised learning within each task, named as meta-learning~\cite{meta1,meta2}. The accumulated meta-knowledge is transferable and the network is able to deal with different tasks. Recently, with the great success of MAML ~\cite{maml}, Huang \etal~\cite{bridge_gap} and Wang \etal~\cite{maml_track} used this fine-tune technique to operate fast adaption for their online tracking task. Inspired by~\cite{attentionrpn,learning_compare}, we aim to learn a transferable deep metric for filtering the distractors by measuring the relationships between proposals and the target via the proposed Relation Detector (RD). Unlike~\cite{attentionrpn}, which assumes the categories of the objects to be detected are given and focuses on detecting the objects based on given categories of support images. However, there is no concept of category in tracking and it is necessary to determine whether the two objects are the same one. 

\section{Method}

 In this section, we first give an overview of our Siamese relation network, then we introduce the Relation Detector in detail, followed by the description of contrast training strategy. Finally, we present our joint Refinement Module.

\subsection{Overview}

 We adopt SiamBAN~\cite{siamban} as our strong baseline which employs ResNet-50 ~\cite{resnet} as the backbone for Siamese network. A cascade structure is designed in our network in order to leverage the multi-level features for more accurate prediction. 
 During inference, the input of the network is a pair of images. One image is the first frame of a sequence cropped to size of $127 \times 127$ and used as template image. The other one is the subsequent frame cropped to $255 \times 255$ and used as search image. After processing them with the backbone we get multi-scale feature maps, which are used for the operation of cross-correlation. Then we utilize Relation Detector (RD), which has learned transferable knowledge based on deep metric, to measure the relationships between target and proposals generated by regression branch. In the end, integrating the outputs of RD and classification results, Refinement Module generates the prediction of the tracking result.
 Compared to a two-stage tracker like SPM~\cite{spm},  which first generates candidates in the same category and then predicts final result within a two-round classification and regression procedure, we only carry out one-round and intend to jointly refine the two branches.
 The tracking pipeline of our method is shown in Figure~\ref{fig:pipeline}.

\begin{figure}
\begin{center}
\setlength{\fboxrule}{0pt}
\setlength{\fboxsep}{0cm}
\fbox{\includegraphics[width=1\linewidth]{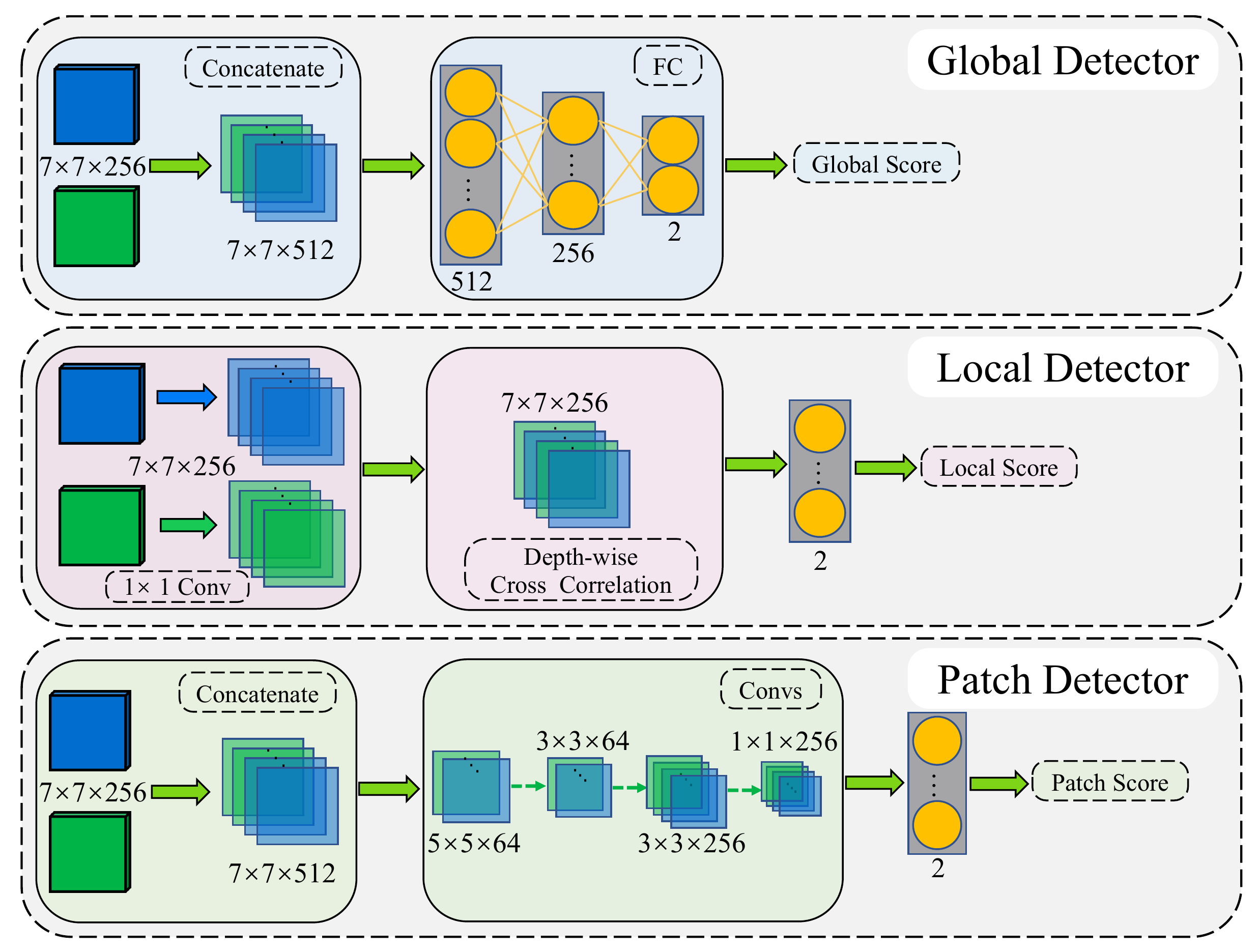}}
\end{center}
  \caption{The architecture of Relation Detector (RD). It consists of three different nonlinear comparators, named as Gloabl Detector, Local Detecor and Patch Detector. We measure the relationships between target and ROI (region of interest) by these detectors which is meta learned to filter the distractors. Each detector finally generates a score to measure the relationship of the input pair. }
\label{fig:rd}
\end{figure}

\subsection{Relation Detector}
\label{section:relation_detector}
The characteristic of the tracking task is that the tracked target can be an arbitrary object, which can be something we have never seen before. 
A naive way is to calculate the difference between proposals and targets by fixed handcraft linear comparator~\cite{metric4,matching} (Cosine distance, Euclidean distance, Mahalanobis distance). However, it can be easily invalid when facing indistinguishable distractors, so we propose a novel Relation Detector (RD) equipped with the adaptive nonlinear comparator that has a substantial discriminative ability to filter the distractors by measuring the similarity relationship with the tracked target. Since training such a detector requires samples of the target in the same sequence, which are usually deficient,  the conventional training strategy can not work well while the way of few-shot learning can overcome this challenge. Therefore we conduct the few-shot learning to train the network so that our Relation Detector can transfer the meta-knowledge gained during massive diverse few-shot task learning. 

The detailed structure of the RD is shown in Figure~\ref{fig:rd}. It consists of three different non-linear learnable comparators, i.e. Global Detector, Local Detector, and Patch Detector. Given the support feature $\mathit{f}_{\mathit{s}}$ and query proposal features $\mathit{f}_{\mathit{q}}$ with the size of $7\times7\times C$, the Global Detector aims to compare their global characteristics through deep embedding, while the Local Detector tries to learn a more detailed comparison in pixel-wise and channel-wise. For the Patch Detector, we design it for learning the relationships between different patches.

However, directly applying the ROI features to the RD may cause the problem of feature misalignments. Thus we introduce a self-attention module~\cite{nonlocal}, which can align the pairs of ROI features before putting them into RD. We conduct the experiment to analyze their performance and effectiveness, which will be further discussed in section~\ref{bf:relation_head}.

\subsection{Contrastive Training Strategy}
\label{section:training_strategy}


    
    Unlike the conventional learning framework, the few-shot learning task has the characteristic of lacking labeled samples in each category~\cite{few-shot}. It aims to construct a classifier to assign a label $\mathit{\hat{y}}$ to each sample $\mathit{\hat{x}}$ in the query set, through some known labeled samples which are considered as support set. When the support set contains $\mathit N$ different categories, and each category has $\mathit K$ labeled samples, we define it as $\mathit N$-way $\mathit K$-shot. In our training, we define two categories, i.e. target and non-target, and conduct our experiments as two-way one-shot learning. 
    
    {\bf Generation of Contrastive Training Samples.} Only matching objects of the same instance is insufficient because the ability to distinguish the different also matters. Thus we exploit the potential relationships of the training samples and construct the training triplet $(\mathit{s}_\mathit{c}, \mathit{q}_\mathit{c}, \mathit{s}_\mathit{n})$, where $\mathit{s}_\mathit{c}$ and $\mathit{s}_\mathit{n}$ are positive support and negative support images, and $\mathit{q}_\mathit{c}$ are query images. $\mathit{s}_\mathit{c}$ and $\mathit{q}_\mathit{c}$ are extracted from the same video while $\mathit{s}_\mathit{n}$ is from different video.
    
    During every single shot, we not only match the objects belonging to the target category but also distinguish distractors in non-target class 
    and the model learns to measure the relationships among different sample combinations generated by the input triplet.
    In detail, we define the ground-truth of positive supports as $\mathit{s}_\mathit{p}$, and use $\mathit{p}_\mathit{p}$ to represent the positive proposal that generated by $\mathit{s}_\mathit{c}$ and $\mathit{q}_\mathit{c}$. Similarly, the ground-truth of negative supports are denoted as $\mathit{n}_\mathit{n}$, and we use $\mathit{p}_\mathit{n}$ for negative proposals that are generated by $\mathit{s}_\mathit{c}$ and $\mathit{q}_\mathit{c}$. Then we combine them to different pairs, as  $(\mathit{s}_\mathit{p}, \mathit{p}_\mathit{p})$, $(\mathit{s}_\mathit{p}, \mathit{p}_\mathit{n})$, $(\mathit{n}_\mathit{n}, \mathit{p}_\mathit{p}/\mathit{p}_\mathit{n})$,  and keep the ratio as 1:2:1. We adopt MSE loss as the loss function and calculate the matching loss on these formed pairs.
    
    {\bf Hard Negative Mining. \label{bf:hnm}} At the early stage of training, applying easy samples can make the model convergent stably. To further enhance the discriminative power of the model, we introduce the hard negative mining technique in the middle and later stage of training.
    In detail, we perform the hard negative mining in two different ways (online and offline). 
    For the online ways,  we first compute the intersection of union (IoU)  between proposals and ground-truth and then remove part of the proposals with IoU value greater than 0.2. Then from the left proposals, we choose the proposals with the highest confidence score as the hard negative samples.
    For the offline ways, we generate an index table for approximate nearest neighbor queries during the pre-processing of large scale datasets including VID~\cite{imagenetvid}, GOT~\cite{got10k} and LaSOT~\cite{lasot}. Then given an image of the tracked object, we use the index table to retrieve nearest neighbors of the tracked object in the embedding space~\cite{siamrcnn} and can find $\mathit{N}$ different hard negative samples.
\subsection{Refinement Module}

\begin{figure}
\begin{center}
\setlength{\fboxrule}{0pt}
\setlength{\fboxsep}{0cm}
\fbox{\includegraphics[width=0.8\linewidth]{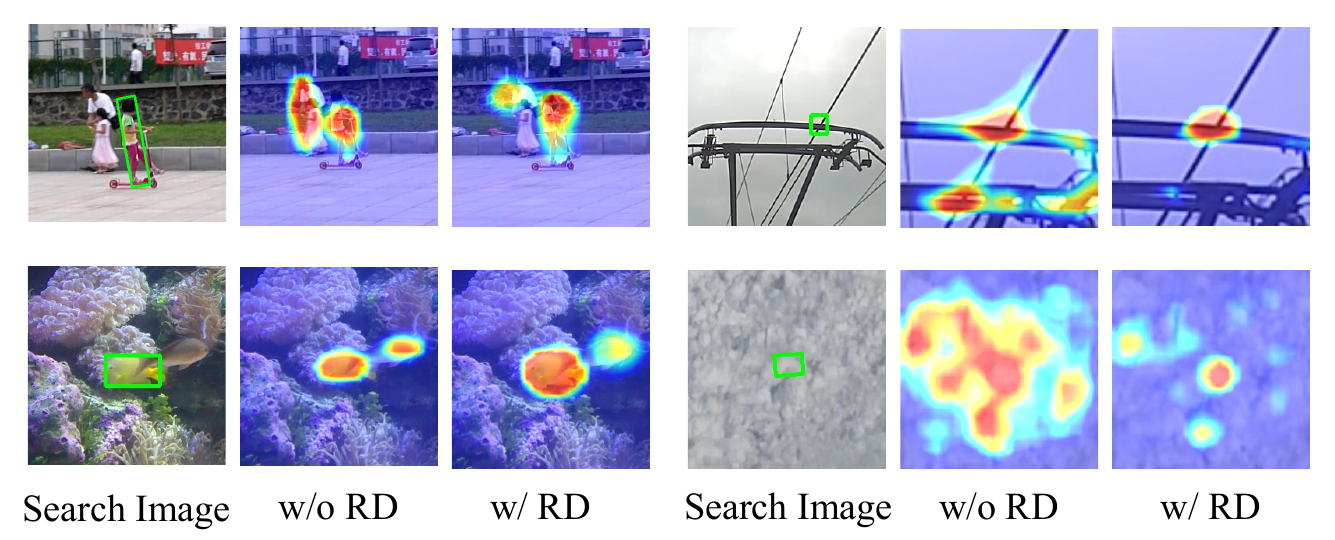}}
\end{center}
  \caption{Visualization of confidence maps. In left part, the $1^{st}$ column represents the search images with ground truth during tracking. The $2^{nd}$ column means the confidence maps without our Relation Detector (RD). The $3^{rd}$ shows the confidence maps that are equipped with Relation Detector (RD), right part is the same. Our Relation Detector (RD) uses the gained meta knowledge to filter the distractors from the target region and advances the discriminability of our tracker.}
\label{fig:heatmap}
\end{figure}

Classification and regression are two pillars in CNN-based trackers, which are usually optimized independently leading to the issue of mismatch between them. To be specific, the box corresponding to the position with the highest classification confidence is not the most accurate, not even the tracked target. We design a refinement module, which effectively integrates the proposed RD into the Siamese framework. At first we utilize the output of the RD and convert it to a matching score sized of $\mathit{25}\times\mathit{25}\times\mathit{1}$. Next, we conduct an element-wise multiplication between the matching score and cross-correlation feature map of the classification branch, which can filter the distractors in the background by suppressing the false positive position. Then we pass the refined correlation feature through a convolution layer to generate a refined classification confidence score. Figure \ref{fig:heatmap} provides some examples of the obtained confidence map. With the refinement module, we combine the information of the regression branch with the classification branch, and jointly operate them to predict the target location so that the problem of mismatch can be alleviated.

\subsection{Ground-truth and Loss function}
    In this section, we illustrate the loss function used training our model. Let $(i,j)$ denote the point in the feature map, and $r_{i,j}$ is the relation score that to be regressed. $(g_{x_c}, g_{y_c})$, $(g_{x_0}, g_{y_0})$, $(g_{x_1}, g_{y_1})$, $(g_{w}, g_{h})$ represent the center position, left-top position, right-bottom position, and size of the target box respectively.
    
{\bf Classification and Regression Branches.}
Followed by Chen \etal~ \cite{siamban}, we use the ellipse figure region to design the label. There are two ellipses namely $E_1$ and $E_2$. The center and axes length of $\mathit{E}_1$ are set to $(g_{x_c},g_{y_c})$ and $(\frac{g_w}{2},\frac{g_h}{2})$ while that of $\mathit{E}_2$ are set to $(g_{x_c},g_{y_c})$ and $(\frac{g_w}{4},\frac{g_h}{4})$. Thus the two ellipse are, 
\begin{equation}
\begin{aligned}
\frac{(p_i - g_{x_c})^2}{(\frac{g_w}{2})^2} + \frac{(p_j - g_{y_c})^2}{(\frac{g_h}{2})^2} = 1 , 
\end{aligned}
\end{equation}

\begin{equation}
\begin{aligned}
\frac{(p_i - g_{x_c})^2}{(\frac{g_w}{4})^2} + \frac{(p_j - g_{y_c})^2}{(\frac{g_h}{4})^2} = 1 , 
\end{aligned}
\end{equation}

For the classification branch, the location $(p_i, p_j)$ falling within the ellipse $\mathit{E}_2$ is defined as positive, and the one which falls outside the ellipse $\mathit{E}_1$ is considered as negative, if $(p_i, p_j)$ falls between the ellipse $\mathit{E}_1$ and $\mathit{E}_2$, we ignore it. Cross-entropy loss is used for calculate the loss function (referred as $\mathcal{L}_{cls}$) of the classification branch. For the regression branch, the positive location $(p_i, p_j)$ is used to regress target box. The ground truth denoted as $d_{i,j}^l$ for the regression branch is defined as,

\begin{equation}
\begin{aligned}
d_{(i,j)}^l = p_i - g_{x_0}, d_{(i,j)}^t = p_j - g_{y_0}, 
\end{aligned}
\end{equation}

\begin{equation}
\begin{aligned}
d_{(i,j)}^r = g_{x_1} - p_i , d_{(i,j)}^b = g_{y_1} - p_j.
\end{aligned}
\end{equation}

We adopt IoU(Intersection over Union) loss for regression, which is defined as,
\begin{equation}
\begin{aligned}
\mathcal{L}_{reg}=1-IoU.
\end{aligned}
\label{eq:iou_loss}
\end{equation}

{\bf Relation Detector.}
The label $y_{i,j}$ for training Relation Detector depends on training pairs mentioned in section \ref{section:training_strategy}.
We adopt MSE loss for our matching score regression, which is formed as, 
\begin{equation}
\begin{aligned}
\mathcal{L}_{matching} = (r_{i,j}- y_{i,j})^2
\end{aligned}
\label{eq:mse_loss}
\end{equation}

Our model is set to train end-to-end and the total loss function is defined as,
\begin{equation}
\begin{aligned}
\mathcal{L}=\lambda_1*\mathcal{L}_{cls}+\lambda_2*\mathcal{L}_{reg}+\lambda_3*\mathcal{L}_{matching},
\end{aligned}
\label{eq:loss}
\end{equation}
we empirically set the $\lambda_1, \lambda_2, \lambda_3$ as $1$ without hyper-parameter searching.

\subsection{Training and Inference}
{\bf Training.} We train our Siamese relation network on large-scale datasets, including ImageNet VID~\cite{imagenetvid}, YouTube-BoundingBoxes~\cite{ytbb}, COCO~\cite{coco}, ImageNet DET~\cite{det}, GOT10k~\cite{got10k} and LaSOT~\cite{lasot}. The training input is an image triplet, including a template patch and a search patch extracted from the same sequence with size of $127 \times 127$ pixels and $255 \times 255$ pixels respectively ~\cite{siamfc}, and a negative search patch which is extracted from another sequence sized as $255 \times 255$. 
We first select two patches from the same sequence in the triplet, and collect at most 16 positive samples and 48 negative samples on it to train the classification branch and the regression branch~\cite{siamrpn, siamrpn++, siamban}. 
Then, the negative search patch in the triplet is used to generate the training input of our Relation Detector, as mentioned in \ref{section:training_strategy}. In addition, we start using online hard negative mining at epoch 5 and epoch 15 for offline. Our entire network can be trained end-to-end and doesn't need any further fine-tuning.

{\bf Inference.} During inference, the patch of the target in the first frame is used as the template and fed it into the backbone to extract the template feature $f_z$. We cache it during tracking to avoid duplicating computation in subsequent tracking. 
Besides, we also generate and cache the ROI feature $f^{roi}_z$ of the template via precise ROI pooling ~\cite{atom}. 
For subsequent frames, we crop the search patch based on the tracking result in the previous frame and extract its feature denoed as $f_x$. Then we perform the prediction in the search region to get the regression map $P_{w\times h\times4}^{reg-all}$ and generate proposals.
Next, features of proposals are cropped and concatenated with the cached target ROI feature 
$f^{roi}_z$. The obtained features are fed in RD to measure the relations between proposals and the target. We convert this relationship to a matching score $s^{matching}_{w\times h\times1}$ and do the element-wise multiplication with the correlation map $f_{cls}^{corr}$ in classification branch. In this way, we fuse the regression result into the classification branch instead of calculating them independently. Then we generate the classification map $P_{w\times h\times2}^{cls-all}$ through the refined correlation map ${f^*}_{cls}^{corr}$. Finally, we can get predicted boxes by $P_{w\times h\times4}^{reg-all}$ and $P_{w\times h\times2}^{cls-all}$.


\section{Experiments}
\begin{table*}
\begin{center}
\resizebox{\textwidth}{!}
{
\begin{tabular}{ccccccccccc}
\hline 

{}        &{\bf ECO}      &{\bf UPDT}           & {\bf SiamRPN}         
            & {\bf LADCF} & {\bf ATOM}           & {\bf SiamRPN++}         
            & {\bf DiMP}  & {\bf SiamBAN}      & {\bf Ocean}           
            &  \multirow{2}*{\bf Ours}            \\ 
            {} &\cite{eco}  &\cite{updt}         &\cite{siamrpn}
            & \cite{ladcf}  &\cite{atom}         &\cite{siamrpn++} 
            & \cite{dimp} &\cite{siamban}         &\cite{ocean}  \\ \hline    
          
EAO$\uparrow$  & $0.281$             & $0.379$               & $0.384$                                         & $0.389$             & $0.401$               & $0.417$              
             & \textcolor{green}{$0.441$}             & \textcolor{blue}{$0.452$}              &\textcolor{red}{$0.470$}
             & \textcolor{red}{$0.470$}           
                  \\
Accuracy$\uparrow$ & $0.484$             & $0.576$               & $0.586$                                 & $0.503$             & $0.590$       &\textcolor{red}{$0.604$}
                 & $0.597$             & \textcolor{green}{$0.597$}               & \textcolor{blue}{$0.598$}
                 & $0.595$    \\
Robustness$\downarrow$ & $0.276$             & $0.184$               & $0.276$                                 & \textcolor{green}{$0.159$}             & $0.204$               & $0.234$  
                     &\textcolor{blue}{$0.152$} & $0.178$               & $0.169$
                     & \textcolor{red}{$0.131$}       \\  \hline 

\end{tabular}
}

\end{center}

\caption{Detail comparisons on VOT2018 with the state-of-the-art in terms of EAO, acccuracy and robustness.  The best three results are shown in \textcolor{red}{red}, \textcolor{blue}{blue} and \textcolor{green}{green} colors, respectively. DiMP is the ResNet-50 version (DiMP-50), and Ocean is the offline version, the same below.}
\label{table:vot2018}
\end{table*}

\begin{table*}
\begin{center}
\resizebox{\textwidth}{!}
{
\begin{tabular}{ccccccccccc}
\hline

{ }         & {\bf MemDTC}        &{\bf SPM}           & {\bf SiamRPN++}         
            & {\bf SiamMask} & {\bf ATOM}           & {\bf DCFST}         
            & {\bf DiMP} & {\bf SiamBAN}      & {\bf Ocean}    
            & \multirow{2}*{\bf Ours}  \\ 
            {} &\cite{memdtc} &\cite{spm}      &\cite{siamrpn++}    &\cite{siammask}  
            &\cite{atom}      &\cite{dcfst}         &\cite{dimp} 
            & \cite{siamban} &\cite{ocean}             \\ \hline
          
EAO$\uparrow$   & $0.228$           & $0.275$               & $0.285$                                               & $0.287$        & $0.292$               & $0.317$              
                & $0.321$            &\textcolor{green}{$0.327$}    & \textcolor{blue}{$0.329$}
             & \textcolor{red}{$0.341$}           
                  \\
Accuracy$\uparrow$   & $0.485$         & $0.577$               & \textcolor{green}{$0.599$}                           & $0.594$          &\textcolor{red}{$0.603$}  & $0.585$        
                 & $0.582$             &\textcolor{blue}{$0.602$}               & $0.590$
                 & $0.593$    \\
Robustness$\downarrow$    & $0.587$          & $0.507$               & $0.482$                                 & $0.461$             & $0.411$            &\textcolor{green}{$0.376$}
                     & \textcolor{blue}{$0.371$}   & $0.396$    & \textcolor{green}{$0.376$}
                     & \textcolor{red}{$0.306$}       \\ \hline

\end{tabular}
}
\end{center}
\caption{Detail comparisons on VOT2019 real-time experiments.}
\label{table:vot2019}
\end{table*}

\subsection{Implementation Details}
\label{section:implement}
We adopt ResNet-50~\cite{resnet} which is pre-trained on ImageNet ~\cite{imagenet_cls} as the backbone, and freeze the weights of the first two layers for stable training. We optimize the model with stochastic gradient descent (SGD) and set the batch size to 28. Our model is trained for 20 epochs, and at the first 5 epochs, we use a warm-up strategy to set the learning rate from $0.001$ to $0.005$. For the last 15 epochs, the learning rate decays exponentially from $0.005$ to $0.00005$. The weights of the backbone are only released in the last 10 epochs and tuned with the learning rate at a one tenth level of current. Weight decay and momentum are set as $0.0001$ and $0.9$. Our approach is implemented in Python using Pytorch with 4 NVIDIA TITAN V GPUs.

\subsection{Comparison with the state-of-the-art}
We evaluate the proposed algorithm against state-of-the-art methods on five tracking
benchmarks.

\begin{figure}
\begin{center}
\setlength{\fboxrule}{0pt}
\setlength{\fboxsep}{0cm}
\fbox{\includegraphics[width=.8\linewidth]{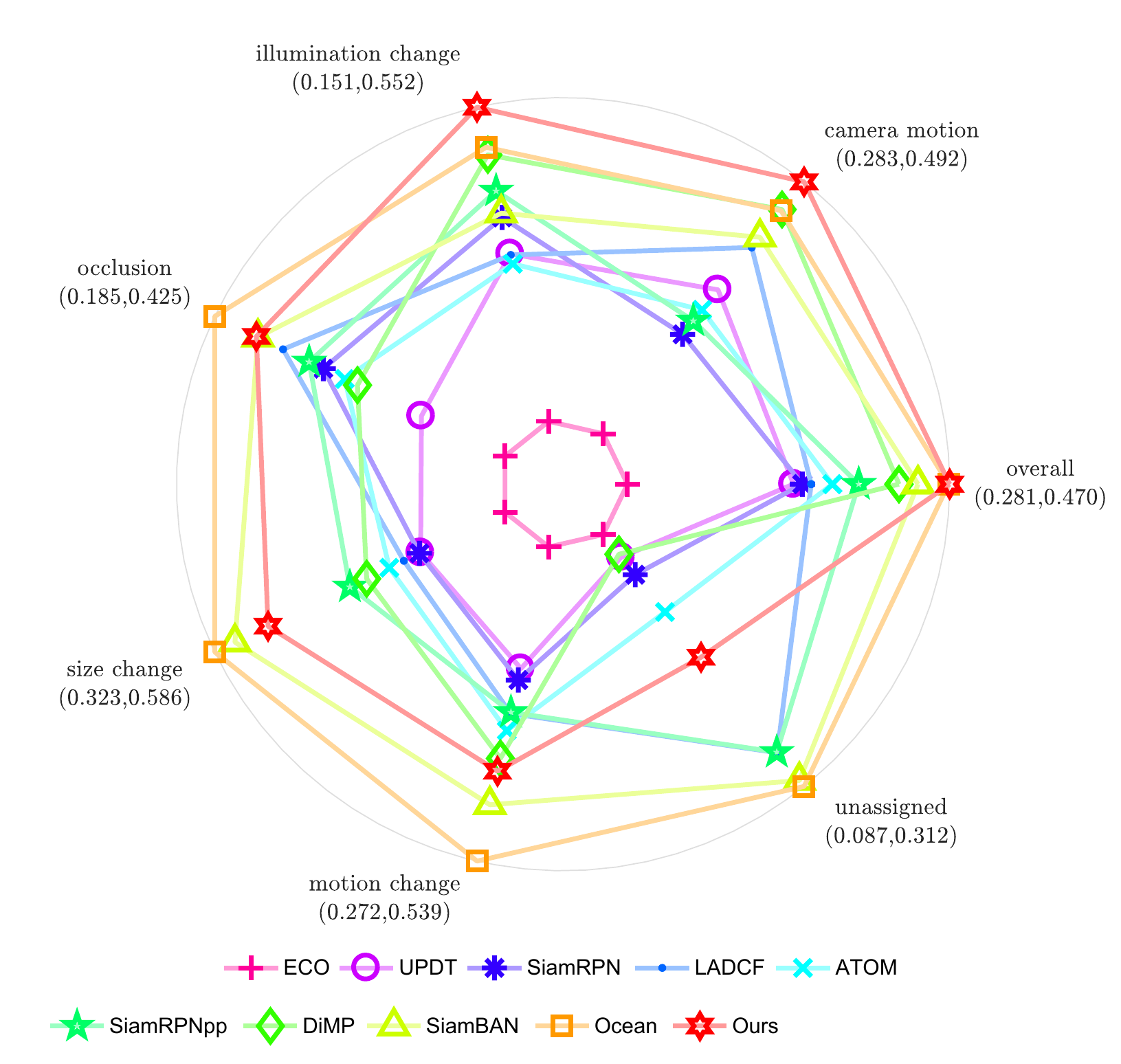}}
\end{center}
  \caption{Comparison of EAO on VOT2018 for the following visual attributes: camera motion, illumination change, occlusion, size change, and motion change. Frames that do not correspond to any of the five attributes are marked as unassigned. The values in parentheses indicate the EAO range of each attribute and overall of the trackers.}
\label{fig:vot2018_arrt}
\end{figure}

\label{section:sota}
{\bf VOT2018}~\cite{vot2018} benchmark consists of 60 sequences with different challenging factors. The overall performance of the tracker is evaluated using the EAO (Expected Average Overlap), which takes both accuracy (average overlap during successful tracking periods) and robustness (failure rate) into consideration. The detailed comparisons with the top-performing trackers are reported in Table \ref{table:vot2018}. Among previous approaches, Ocean~\cite{ocean} (offline version) achieves the best EAO value and SiamRPN++~\cite{siamrpn++} gets the highest accuracy. These two methods are both based on the Siamese network. Compared with DiMP~\cite{dimp}, our model achieves a performance gain of $2.9\%$. When compared to Ocean, our model achieves the same EAO value that is the highest value among the proposed state-of-the-art trackers. In comparison with the baseline tracker SiamBAN~\cite{siamban}, we have a substantial improvement of $4.7\%$ in robustness. These results prove that our Siamese Relation Network has learned the strong ability to filter the distractors as expected. Furthermore, we conduct a comparison with the sate-of-the-art trackers in term of EAO on different visual attributes and the result is shown in Figure~\ref{fig:vot2018_arrt}. 
  On attributes of camera motion and illumination change our tracker ranks first, and ranks second and third on attributes of occlusion and size change. This shows our tracker are capable to tackle above challenges.

{\bf VOT2019}~\cite{vot2019result} public dataset is one of the most recent datasets for evaluating single object trackers. Compared with VOT2018, the video sequences in VOT2019 have a 20\% difference and more challenging videos which contain fast motion and similar distractors. Table \ref{table:vot2019} shows that our tracker achieves the optimal performance in EAO, which is $1.2\%$ higher than that of Ocean which ranked at second place. In addition, in comparison with the baseline SiamBAN, we achieve a $9.0\%$ improvement of robustness, which proves the strong discriminability of our tracker.

\begin{figure}
\begin{center}
\setlength{\fboxrule}{0pt}
\setlength{\fboxsep}{0cm}
\fbox{\includegraphics[width=.8\linewidth]{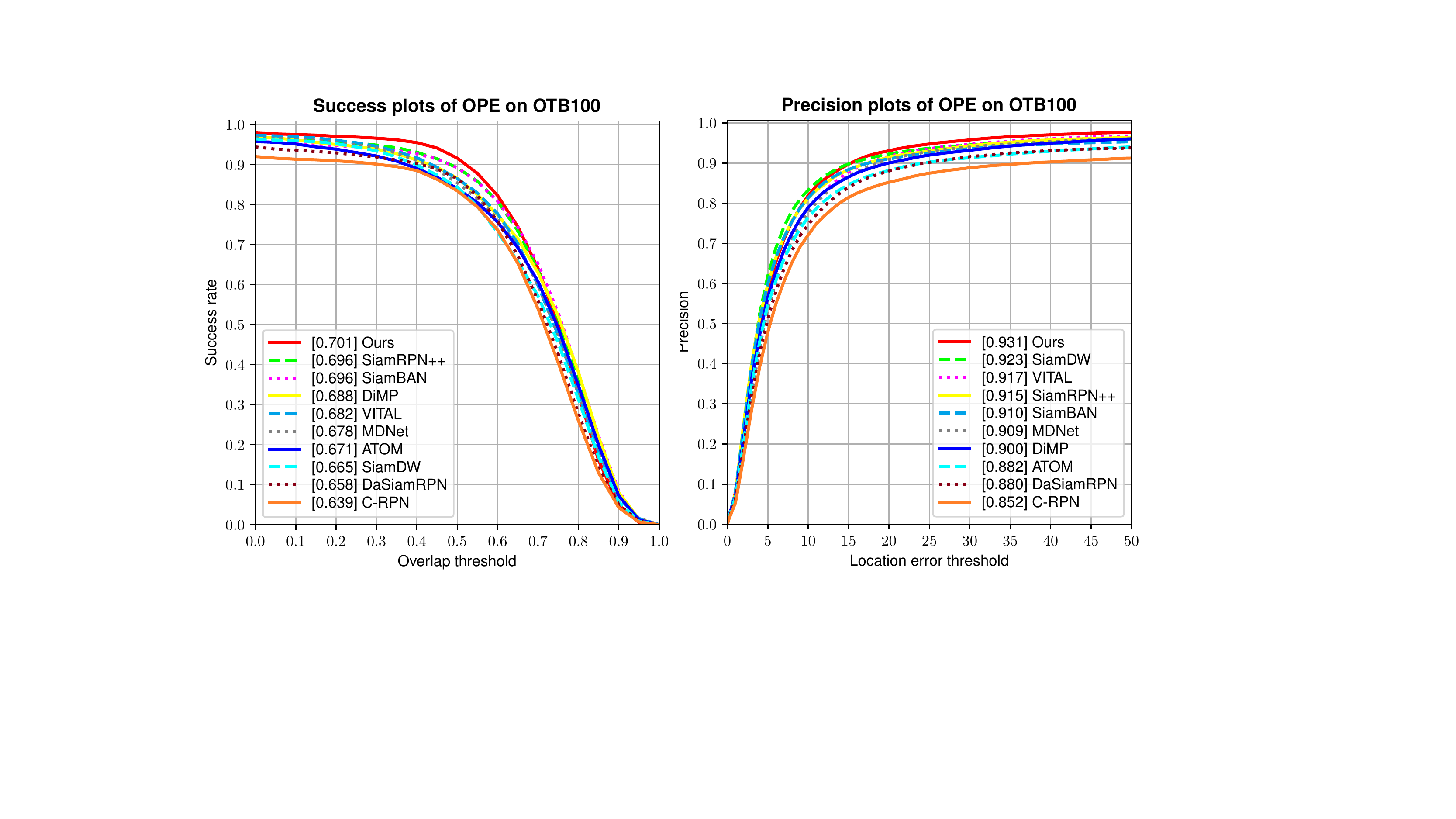}}
\end{center}
  \caption{Success and precision plots on OTB100.}
\label{fig:otb}
\end{figure}

{\bf OTB100}~\cite{otb2015} is one of the most widely used benchmarks for visual object tracking and consists of 100 well-annotated video sequences. Different from VOT settings, it has a regulation known as one-pass evaluation (OPE) and evaluates the tracker on two basic indications: a precision score and an area under curve (AUC) of success plot. Compared with the top-performing trackers as shown in Figure \ref{fig:otb}, our model has a good performance in challenging videos and achieves the leading position among sate-of-the-art trackers.

\begin{figure}
\begin{center}
\setlength{\fboxrule}{0pt}
\setlength{\fboxsep}{0cm}
\fbox{\includegraphics[width=.8\linewidth]{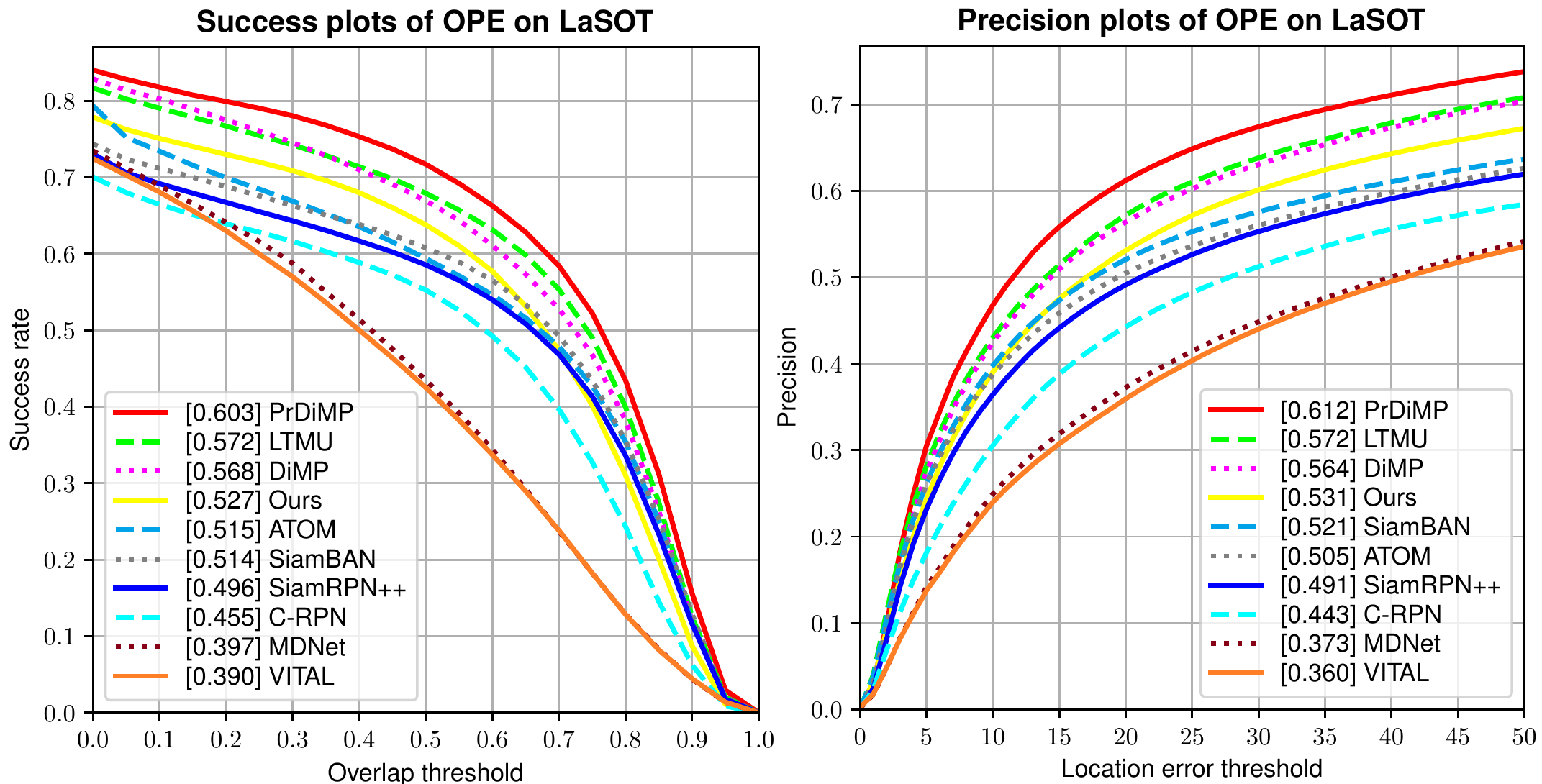}}
\end{center}
  \caption{Success and precision plots on LaSOT.}
\label{fig:lasot}
\end{figure}

{\bf LaSOT}~\cite{lasot}
is a large-scale dataset with 1400 sequences in total. The video sequences in LaSOT have an average sequence length of more than 2500 frames, which is much longer than previous datasets. We conduct one-pass evaluation with success and precision scores to evaluate our tracker. The result is shown in Figure~\ref{fig:lasot}. Compared with nine methods~\cite{prdimp, dimp, siamban, atom, siamrpn++, crpn, mdnet, vital, ltmu}, our approach achieves the fourth place in terms of AUC and precision without any long-term strategies, which is $1.3\%$ and $1.0\%$ higher than baseline.

\begin{figure}
\begin{center}
\setlength{\fboxrule}{0pt}
\setlength{\fboxsep}{0cm}
\fbox{\includegraphics[width=.8\linewidth]{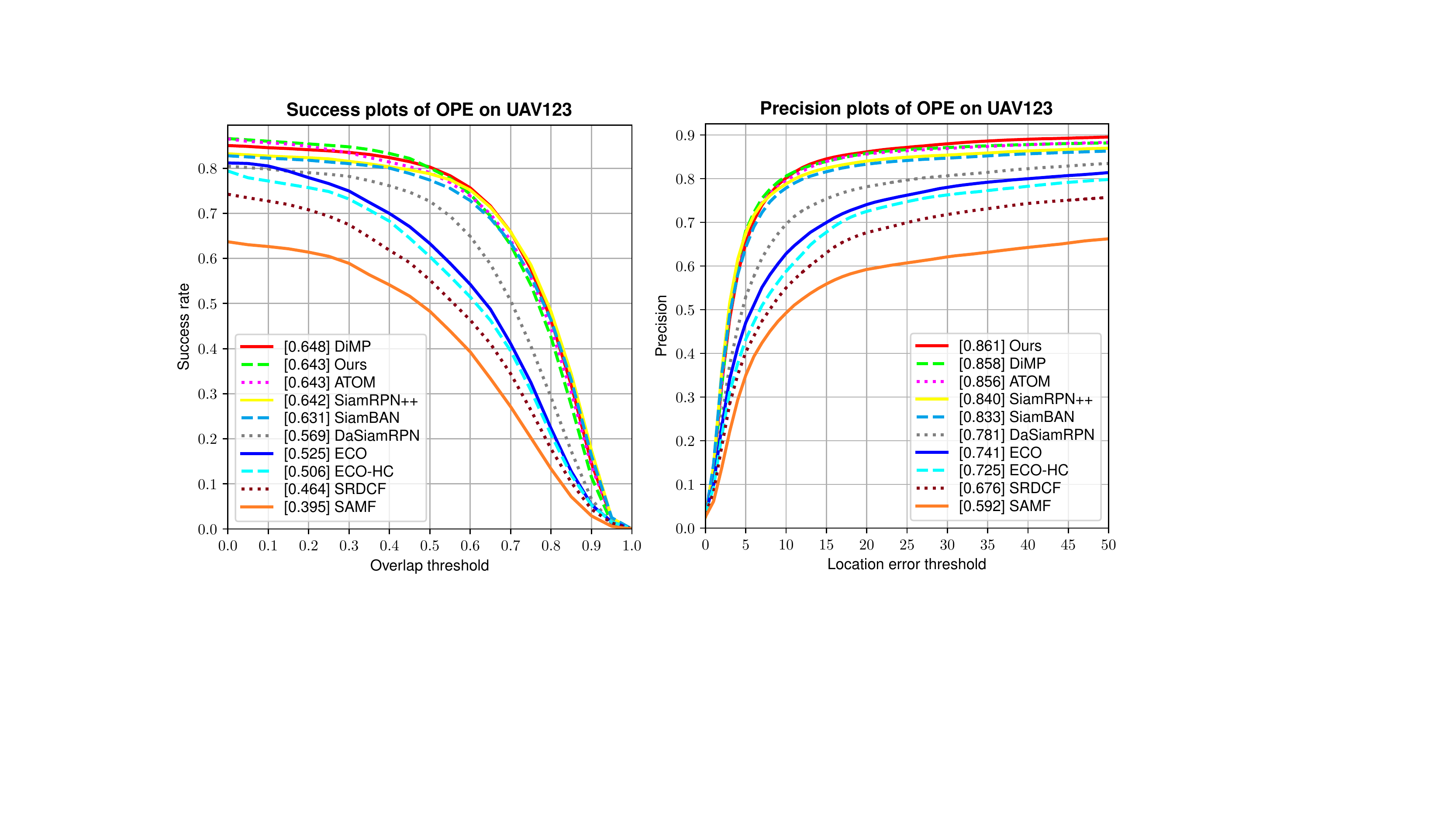}}
\end{center}
  \caption{Success and precision plots on UAV123.}
\label{fig:uav123}
\end{figure}

{\bf UAV123}~\cite{uav123} is an aerial video benchmark that contains 123 sequences captured from low-altitude UAVs. Different from other benchmarks, the perspective of UAV123 is from top to bottom, and the size of the target is relatively small. Compared with 9 state-of-art real-time trackers~\cite{dimp, atom, siamrpn++, siamban, dasiamrpn, eco, srdcf, samf}, our tracker ranks second in AUC and first in precision, see in Figure~\ref{fig:uav123}.

\begin{figure}
\begin{center}
\setlength{\fboxrule}{0pt}
\setlength{\fboxsep}{0cm}
\fbox{\includegraphics[width=.8\linewidth]{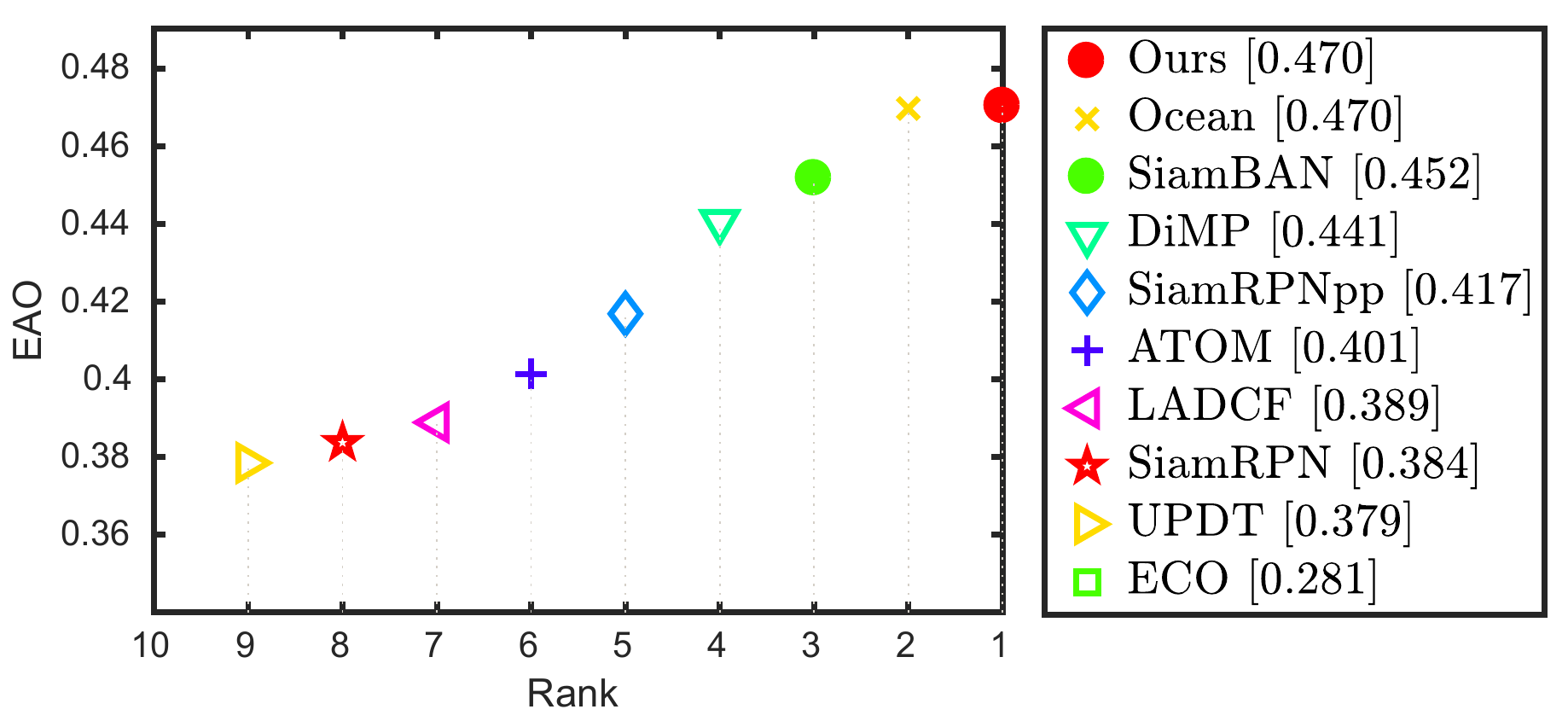}}
\end{center}
  \caption{Expected averaged overlap performance on VOT2018.}
\label{fig:vot2018}
\end{figure}

\begin{figure}
\begin{center}
\setlength{\fboxrule}{0pt}
\setlength{\fboxsep}{0cm}
\fbox{\includegraphics[width=.8\linewidth]{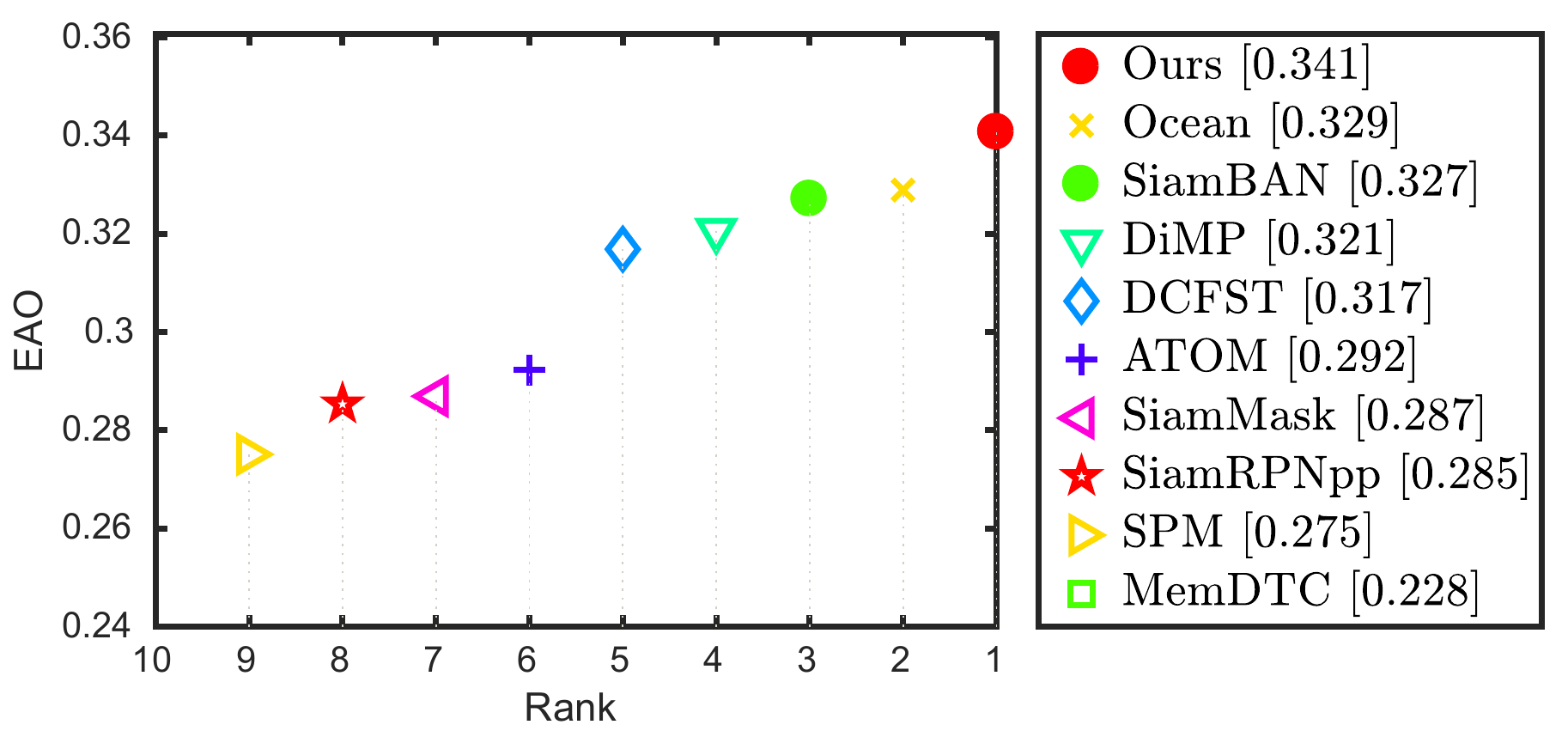}}
\end{center}
  \caption{Expected averaged overlap performance on VOT2019.}
\label{fig:vot2019}
\end{figure}


\subsection{Ablation Study}
\label{section:ablation}
{\bf Discussion on Multi-level Prediction and Relation Detection.}
The multi-level feature contains different information on the target, in order to fully exploit the potential of them, we investigate the impact of layer-wise aggregation. The Relation Detector performs the relation detection on a different level as well to capture the multiple relationships between target and proposals. From the results shown in Table \ref{table:ablation} we can find that when only utilizing a single-layer feature, $conv4$ performs best. For different combinations of two-level features, we can see that the combination of $conv4$ and $conv5$ is the best. After aggregating three-level features, our model gets the highest score on AUC.

{\bf Discussion on Different Relation Head.}
{\label{bf:relation_head}}
We test the performance of the three kinds of relation detectors known as Global Detector, Local Detector, and Patch Detector, which has been mentioned in 
\ref{section:relation_detector}. Trying to find out the relation we need most, we conduct the experiments following the 2-way 1-shot setting on OTB100.
The results of the ablation study are shown in Table \ref{table:ablation}. For adopting a single head, the Local Detector gets the best score on AUC. We infer that it can compare the ROI feature most in detail. Combining two types of relation heads achievs better performance than using the single one. The best performance is achieved when utilizing all types of relations heads. It seems that these three kinds of heads are complementary with each other for distinguishing the target from background clutters.

\begin{table}
\begin{center}
\begin{tabular}{ccc|ccc|c}
\hline \hline
C3              & C4            & C5     &Global  & Local     & Patch & AUC $\uparrow$ \\ \hline
\checkmark      & {}            &{}      &\checkmark   &\checkmark   &\checkmark    & $0.676$ \\
{}              & \checkmark    & {}     &\checkmark   &\checkmark   &\checkmark    & $0.686$ \\
{}              & {}            &\checkmark &\checkmark  &\checkmark &\checkmark    & $0.669$ \\
\checkmark      &\checkmark     &{}      &\checkmark   &\checkmark   &\checkmark    & $0.691$ \\
\checkmark      & {}            &\checkmark &\checkmark  &\checkmark &\checkmark    &$0.684$ \\
{}      & \checkmark  &\checkmark         &\checkmark   &\checkmark   &\checkmark           & $0.693$ \\ \hline
\checkmark   &\checkmark    &\checkmark   &\checkmark   & {}            & {}  
& $0.678$ \\ 
\checkmark   &\checkmark    &\checkmark   & {}          &\checkmark     & {}  
& $0.686$ \\
\checkmark   &\checkmark    &\checkmark   & {}          & {}        &\checkmark     &$0.672$ 
\\
\checkmark   &\checkmark    &\checkmark   &\checkmark   &\checkmark    & {}         &$0.696$ 
\\
\checkmark   &\checkmark    &\checkmark   &\checkmark   & {}            &\checkmark & $0.683$
\\
\checkmark   &\checkmark    &\checkmark   & {}          &\checkmark    &\checkmark & $0.693$
\\ \hline
\checkmark   &\checkmark    &\checkmark   &\checkmark   &\checkmark    &\checkmark        &{\bf 0.701} \\ \hline \hline

\end{tabular}
\end{center}
\caption{Quantitative comparison results of our trackers with different level feature combinations and different relation heads combinations on OTB100. C3, C4, C5 represent $\mathit{conv}$3, $\mathit{conv}$4, $\mathit{conv}$5. Global, Local, Patch represent three types of relation heads proposed in \ref{section:relation_detector}.}
\label{table:ablation}
\end{table}

{\bf Discussion on Contrastive Training Strategy and Hard Negative Mining.}
During the few-shot training phase, in order to improve the discriminability of our model, we perform a contrastive training strategy described in section~\ref{section:training_strategy}. when conduction experiments on OTB100, we find that if using the naive strategy (one-way one-shot learning), the model achieves $0.688$ in AUC, which drops $1.3\%$ compared to the contrastive way. It seems that our contrastive training strategy not only learns the similarity between target and proposals but also can distinguish the difference between them. Moreover, our proposed hard negative mining technique enhances $0.3\%$ in AUC. It maybe because that adding 
hard negative samples in time will improve the robustness of the mode learned from massive simple samples.

\section{Conclusions}
In the paper, we propose a simple yet effective model called Relation Detector (RD) equipped with contrastive training strategy, which is meta trained to gain the ability to learn to filter the distractors from the target region by measuring the relationships between them. Moreover, we design a refinement module to jointly operate the classification and regression process to localize the target, which can alleviate the mismatch of these two branches to get more confident and accurate results. Extensive experiments are conducted on five popular benchmarks and our method obtains the state-of-the-art results with real-time running speed. 
\section*{Acknowledgments}
This work was supported by the National Natural Science Foundation of China (No. 61972167, 61802135), the Project of Guangxi Science and Technology (No. 2020AC19194), the Guangxi “Bagui Scholar” Teams for Innovation and Research Project, the Guangxi Collaborative Innovation Center of Multi-source Information Integration and Intelligent Processing, the Guangxi Talent Highland Project of Big Data Intelligence and Application, and the Open Project Program of the National Laboratory of Pat- tern Recognition (NLPR) (No. 202000012).

{\small
\bibliographystyle{ieee_fullname}
\bibliography{paper}

\begin{thebibliography}{10}\itemsep=-1pt

\bibitem{siamfc}
Luca Bertinetto, Jack Valmadre, Jo{\~{a}}o~F. Henriques, Andrea Vedaldi, and
  Philip H.~S. Torr.
\newblock Fully-convolutional siamese networks for object tracking.
\newblock In {\em {ECCV} Workshops {(2)}}, volume 9914 of {\em Lecture Notes in
  Computer Science}, pages 850--865, 2016.

\bibitem{dimp}
Goutam Bhat, Martin Danelljan, and Luc~Van Gool.
\newblock Learning discriminative model prediction for tracking.
\newblock In {\em {ICCV}}, pages 6181--6190. {IEEE}, 2019.

\bibitem{updt}
Goutam Bhat, Joakim Johnander, Martin Danelljan, Fahad~Shahbaz Khan, and
  Michael Felsberg.
\newblock Unveiling the power of deep tracking.
\newblock In {\em {ECCV} {(2)}}, volume 11206 of {\em Lecture Notes in Computer
  Science}, pages 493--509. Springer, 2018.

\bibitem{siamban}
Zedu Chen, Bineng Zhong, Guorong Li, Shengping Zhang, and Rongrong Ji.
\newblock Siamese box adaptive network for visual tracking.
\newblock In {\em {CVPR}}, pages 6667--6676. {IEEE}, 2020.

\bibitem{ltmu}
Kenan Dai, Yunhua Zhang, Dong Wang, Jianhua Li, Huchuan Lu, and Xiaoyun Yang.
\newblock High-performance long-term tracking with meta-updater.
\newblock In {\em {CVPR}}, pages 6297--6306. {IEEE}, 2020.

\bibitem{atom}
Martin Danelljan, Goutam Bhat, and Fahad~Shahbaz Khan.
\newblock {ATOM:} accurate tracking by overlap maximization.
\newblock In {\em {CVPR}}, pages 4660--4669. Computer Vision Foundation /
  {IEEE}, 2019.

\bibitem{eco}
Martin Danelljan, Goutam Bhat, Fahad~Shahbaz Khan, and Michael Felsberg.
\newblock {ECO:} efficient convolution operators for tracking.
\newblock In {\em {CVPR}}, pages 6931--6939. {IEEE} Computer Society, 2017.

\bibitem{prdimp}
Martin Danelljan, Luc~Van Gool, and Radu Timofte.
\newblock Probabilistic regression for visual tracking.
\newblock In {\em {CVPR}}, pages 7181--7190. {IEEE}, 2020.

\bibitem{srdcf}
Martin Danelljan, Gustav H{\"{a}}ger, Fahad~Shahbaz Khan, and Michael Felsberg.
\newblock Learning spatially regularized correlation filters for visual
  tracking.
\newblock In {\em {ICCV}}, pages 4310--4318. {IEEE} Computer Society, 2015.

\bibitem{meta2}
Jennifer~G. Dy and Andreas Krause.
\newblock Proceedings of the 35th international conference on machine learning,
  {ICML} 2018, stockholmsm{\"{a}}ssan, stockholm, sweden, july 10-15, 2018.
\newblock volume~80 of {\em Proceedings of Machine Learning Research}. {PMLR},
  2018.

\bibitem{lasot}
Heng Fan, Liting Lin, and Fan Yang.
\newblock Lasot: {A} high-quality benchmark for large-scale single object
  tracking.
\newblock In {\em {CVPR}}, pages 5374--5383. Computer Vision Foundation /
  {IEEE}, 2019.

\bibitem{crpn}
Heng Fan and Haibin Ling.
\newblock Siamese cascaded region proposal networks for real-time visual
  tracking.
\newblock In {\em {CVPR}}, pages 7952--7961. Computer Vision Foundation /
  {IEEE}, 2019.

\bibitem{attentionrpn}
Qi Fan, Wei Zhuo, Chi{-}Keung Tang, and Yu{-}Wing Tai.
\newblock Few-shot object detection with attention-rpn and multi-relation
  detector.
\newblock In {\em {CVPR}}, pages 4012--4021. {IEEE}, 2020.

\bibitem{maml}
Chelsea Finn, Pieter Abbeel, and Sergey Levine.
\newblock Model-agnostic meta-learning for fast adaptation of deep networks.
\newblock In {\em {ICML}}, volume~70 of {\em Proceedings of Machine Learning
  Research}, pages 1126--1135. {PMLR}, 2017.

\bibitem{dsiam}
Qing Guo, Wei Feng, Ce Zhou, Rui Huang, Liang Wan, and Song Wang.
\newblock Learning dynamic siamese network for visual object tracking.
\newblock In {\em {ICCV}}, pages 1781--1789. {IEEE} Computer Society, 2017.

\bibitem{resnet}
Kaiming He, Xiangyu Zhang, Shaoqing Ren, and Jian Sun.
\newblock Deep residual learning for image recognition.
\newblock In {\em {CVPR}}, pages 770--778. {IEEE} Computer Society, 2016.

\bibitem{kcf}
Jo{\~{a}}o~F. Henriques, Rui Caseiro, Pedro Martins, and Jorge Batista.
\newblock High-speed tracking with kernelized correlation filters.
\newblock {\em {IEEE} Trans. Pattern Anal. Mach. Intell.}, 37(3):583--596,
  2015.

\bibitem{got10k}
Lianghua Huang, Xin Zhao, and Kaiqi Huang.
\newblock Got-10k: {A} large high-diversity benchmark for generic object
  tracking in the wild.
\newblock {\em CoRR}, abs/1810.11981, 2018.

\bibitem{bridge_gap}
Lianghua Huang, Xin Zhao, and Kaiqi Huang.
\newblock Bridging the gap between detection and tracking: {A} unified
  approach.
\newblock In {\em {ICCV}}, pages 3998--4008. {IEEE}, 2019.

\bibitem{mal}
Wei Ke, Tianliang Zhang, Zeyi Huang, Qixiang Ye, Jianzhuang Liu, and Dong
  Huang.
\newblock Multiple anchor learning for visual object detection.
\newblock In {\em {CVPR}}, pages 10203--10212. {IEEE}, 2020.

\bibitem{vot2018}
Matej Kristan, Ales Leonardis, and Jiri Matas.
\newblock The sixth visual object tracking {VOT2018} challenge results.
\newblock In {\em {ECCV} Workshops {(1)}}, volume 11129 of {\em Lecture Notes
  in Computer Science}, pages 3--53. Springer, 2018.

\bibitem{cnn}
Alex Krizhevsky, Ilya Sutskever, and Geoffrey~E. Hinton.
\newblock Imagenet classification with deep convolutional neural networks.
\newblock In {\em {NIPS}}, pages 1106--1114, 2012.

\bibitem{imagenet_cls}
Alex Krizhevsky, Ilya Sutskever, and Geoffrey~E. Hinton.
\newblock Imagenet classification with deep convolutional neural networks.
\newblock In {\em {NIPS}}, pages 1106--1114, 2012.

\bibitem{siamrpn++}
Bo Li, Wei Wu, Qiang Wang, Fangyi Zhang, Junliang Xing, and Junjie Yan.
\newblock Siamrpn++: Evolution of siamese visual tracking with very deep
  networks.
\newblock In {\em {CVPR}}, pages 4282--4291. Computer Vision Foundation /
  {IEEE}, 2019.

\bibitem{siamrpn}
Bo Li, Junjie Yan, Wei Wu, Zheng Zhu, and Xiaolin Hu.
\newblock High performance visual tracking with siamese region proposal
  network.
\newblock In {\em {CVPR}}, pages 8971--8980. {IEEE} Computer Society, 2018.

\bibitem{few-shot1}
Fei{-}Fei Li, Robert Fergus, and Pietro Perona.
\newblock One-shot learning of object categories.
\newblock {\em {IEEE} Trans. Pattern Anal. Mach. Intell.}, 28(4):594--611,
  2006.

\bibitem{1survey}
Xi Li, Weiming Hu, Chunhua Shen, Zhongfei Zhang, Anthony~R. Dick, and Anton
  van~den Hengel.
\newblock A survey of appearance models in visual object tracking.
\newblock {\em {ACM} Trans. Intell. Syst. Technol.}, 4(4):58:1--58:48, 2013.

\bibitem{autodrive}
Xi Li, Weiming Hu, Chunhua Shen, Zhongfei Zhang, Anthony~R. Dick, and Anton
  van~den Hengel.
\newblock A survey of appearance models in visual object tracking.
\newblock {\em {ACM} Trans. Intell. Syst. Technol.}, 4(4):58:1--58:48, 2013.

\bibitem{samf}
Yang Li and Jianke Zhu.
\newblock A scale adaptive kernel correlation filter tracker with feature
  integration.
\newblock In {\em {ECCV} Workshops {(2)}}, volume 8926 of {\em Lecture Notes in
  Computer Science}, pages 254--265. Springer, 2014.

\bibitem{coco}
Tsung{-}Yi Lin, Michael Maire, Serge~J. Belongie, James Hays, Pietro Perona,
  Deva Ramanan, Piotr Doll{\'{a}}r, and C.~Lawrence Zitnick.
\newblock Microsoft {COCO:} common objects in context.
\newblock In {\em {ECCV} {(5)}}, volume 8693 of {\em Lecture Notes in Computer
  Science}, pages 740--755. Springer, 2014.

\bibitem{human_computer}
Liwei Liu, Junliang Xing, Haizhou Ai, and Xiang Ruan.
\newblock Hand posture recognition using finger geometric feature.
\newblock In {\em {ICPR}}, pages 565--568. {IEEE} Computer Society, 2012.

\bibitem{vot2019result}
Kristan M, Matas J, and Leonardis A.
\newblock The seventh visual object tracking {VOT2019} challenge results.
\newblock In {\em {ICCV} Workshops}, pages 2206--2241. {IEEE}, 2019.

\bibitem{uav123}
Matthias Mueller, Neil Smith, and Bernard Ghanem.
\newblock A benchmark and simulator for {UAV} tracking.
\newblock In {\em {ECCV} {(1)}}, volume 9905 of {\em Lecture Notes in Computer
  Science}, pages 445--461. Springer, 2016.

\bibitem{mdnet}
Hyeonseob Nam and Bohyung Han.
\newblock Learning multi-domain convolutional neural networks for visual
  tracking.
\newblock In {\em {CVPR}}, pages 4293--4302. {IEEE} Computer Society, 2016.

\bibitem{metric1}
Boris~N. Oreshkin, Pau~Rodr{\'{\i}}guez L{\'{o}}pez, and Alexandre Lacoste.
\newblock {TADAM:} task dependent adaptive metric for improved few-shot
  learning.
\newblock In {\em NeurIPS}, pages 719--729, 2018.

\bibitem{meta1}
Doina Precup and Yee~Whye Teh.
\newblock Proceedings of the 34th international conference on machine learning,
  {ICML} 2017, sydney, nsw, australia, 6-11 august 2017.
\newblock volume~70 of {\em Proceedings of Machine Learning Research}. {PMLR},
  2017.

\bibitem{ytbb}
Esteban Real, Jonathon Shlens, Stefano Mazzocchi, Xin Pan, and Vincent
  Vanhoucke.
\newblock Youtube-boundingboxes: {A} large high-precision human-annotated data
  set for object detection in video.
\newblock In {\em {CVPR}}, pages 7464--7473. {IEEE} Computer Society, 2017.

\bibitem{faster_rcnn}
Shaoqing Ren, Kaiming He, Ross~B. Girshick, and Jian Sun.
\newblock Faster {R-CNN:} towards real-time object detection with region
  proposal networks.
\newblock In {\em {NIPS}}, pages 91--99, 2015.

\bibitem{imagenetvid}
Olga Russakovsky, Jia Deng, and Hao Su.
\newblock Imagenet large scale visual recognition challenge.
\newblock {\em Int. J. Comput. Vis.}, 115(3):211--252, 2015.

\bibitem{det}
Olga Russakovsky, Jia Deng, and Hao Su.
\newblock Imagenet large scale visual recognition challenge.
\newblock {\em Int. J. Comput. Vis.}, 115(3):211--252, 2015.

\bibitem{DBLP:rnn1}
Adam Santoro, Sergey Bartunov, Matthew Botvinick, Daan Wierstra, and Timothy~P.
  Lillicrap.
\newblock Meta-learning with memory-augmented neural networks.
\newblock In {\em {ICML}}, volume~48 of {\em {JMLR} Workshop and Conference
  Proceedings}, pages 1842--1850. JMLR.org, 2016.

\bibitem{metric4}
Jake Snell, Kevin Swersky, and Richard~S. Zemel.
\newblock Prototypical networks for few-shot learning.
\newblock In {\em {NIPS}}, pages 4077--4087, 2017.

\bibitem{vital}
Yibing Song, Chao Ma, and Xiaohe Wu.
\newblock {VITAL:} visual tracking via adversarial learning.
\newblock In {\em {CVPR}}, pages 8990--8999. {IEEE} Computer Society, 2018.

\bibitem{learning_compare}
Flood Sung, Yongxin Yang, Li Zhang, Tao Xiang, Philip H.~S. Torr, and
  Timothy~M. Hospedales.
\newblock Learning to compare: Relation network for few-shot learning.
\newblock In {\em {CVPR}}, pages 1199--1208. {IEEE} Computer Society, 2018.

\bibitem{few-shot}
Sebastian Thrun.
\newblock Is learning the n-th thing any easier than learning the first?
\newblock In {\em {NIPS}}, pages 640--646. {MIT} Press, 1995.

\bibitem{fcos}
Zhi Tian, Chunhua Shen, Hao Chen, and Tong He.
\newblock {FCOS:} fully convolutional one-stage object detection.
\newblock In {\em {ICCV}}, pages 9626--9635. {IEEE}, 2019.

\bibitem{cfnet}
Jack Valmadre, Luca Bertinetto, Jo{\~{a}}o~F. Henriques, Andrea Vedaldi, and
  Philip H.~S. Torr.
\newblock End-to-end representation learning for correlation filter based
  tracking.
\newblock In {\em {CVPR}}, pages 5000--5008. {IEEE} Computer Society, 2017.

\bibitem{matching}
Oriol Vinyals, Charles Blundell, Tim Lillicrap, Koray Kavukcuoglu, and Daan
  Wierstra.
\newblock Matching networks for one shot learning.
\newblock In {\em {NIPS}}, pages 3630--3638, 2016.

\bibitem{siamrcnn}
Paul Voigtlaender, Jonathon Luiten, and Philip H.~S. Torr.
\newblock Siam {R-CNN:} visual tracking by re-detection.
\newblock In {\em {CVPR}}, pages 6577--6587. {IEEE}, 2020.

\bibitem{maml_track}
Guangting Wang, Chong Luo, Xiaoyan Sun, Zhiwei Xiong, and Wenjun Zeng.
\newblock Tracking by instance detection: {A} meta-learning approach.
\newblock In {\em {CVPR}}, pages 6287--6296. {IEEE}, 2020.

\bibitem{spm}
Guangting Wang, Chong Luo, and Zhiwei Xiong.
\newblock Spm-tracker: Series-parallel matching for real-time visual object
  tracking.
\newblock In {\em {CVPR}}, pages 3643--3652. Computer Vision Foundation /
  {IEEE}, 2019.

\bibitem{siammask}
Qiang Wang, Li Zhang, and Luca Bertinetto.
\newblock Fast online object tracking and segmentation: {A} unifying approach.
\newblock In {\em {CVPR}}, pages 1328--1338. Computer Vision Foundation /
  {IEEE}, 2019.

\bibitem{nonlocal}
Xiaolong Wang, Ross~B. Girshick, Abhinav Gupta, and Kaiming He.
\newblock Non-local neural networks.
\newblock In {\em {CVPR}}, pages 7794--7803. {IEEE} Computer Society, 2018.

\bibitem{otb2015}
Yi Wu, Jongwoo Lim, and Ming{-}Hsuan Yang.
\newblock Online object tracking: {A} benchmark.
\newblock In {\em {CVPR}}, pages 2411--2418. {IEEE} Computer Society, 2013.

\bibitem{challenge}
Yi Wu, Jongwoo Lim, and Ming{-}Hsuan Yang.
\newblock Object tracking benchmark.
\newblock {\em {IEEE} Trans. Pattern Anal. Mach. Intell.}, 37(9):1834--1848,
  2015.

\bibitem{surveillance}
Ruiyue Xu, Yepeng Guan, and Yizhen Huang.
\newblock Multiple human detection and tracking based on head detection for
  real-time video surveillance.
\newblock {\em Multim. Tools Appl.}, 74(3):729--742, 2015.

\bibitem{ladcf}
Tianyang Xu, Zhen{-}Hua Feng, Xiao{-}Jun Wu, and Josef Kittler.
\newblock Learning adaptive discriminative correlation filters via temporal
  consistency preserving spatial feature selection for robust visual object
  tracking.
\newblock {\em {IEEE} Trans. Image Process.}, 28(11):5596--5609, 2019.

\bibitem{memdtc}
Tianyu Yang and Antoni~B. Chan.
\newblock Learning dynamic memory networks for object tracking.
\newblock In {\em {ECCV} {(9)}}, volume 11213 of {\em Lecture Notes in Computer
  Science}, pages 153--169. Springer, 2018.

\bibitem{updatenet}
Lichao Zhang, Abel Gonzalez{-}Garcia, Joost van~de Weijer, Martin Danelljan,
  and Fahad~Shahbaz Khan.
\newblock Learning the model update for siamese trackers.
\newblock In {\em {ICCV}}, pages 4009--4018. {IEEE}, 2019.

\bibitem{siamdw}
Zhipeng Zhang and Houwen Peng.
\newblock Deeper and wider siamese networks for real-time visual tracking.
\newblock In {\em {CVPR}}, pages 4591--4600. Computer Vision Foundation /
  {IEEE}, 2019.

\bibitem{ocean}
Zhipeng Zhang, Houwen Peng, and Jianlong Fu.
\newblock Ocean: Object-aware anchor-free tracking.
\newblock In {\em {ECCV} {(21)}}, volume 12366 of {\em Lecture Notes in
  Computer Science}, pages 771--787. Springer, 2020.

\bibitem{dcfst}
Linyu Zheng, Ming Tang, Yingying Chen, Jinqiao Wang, and Hanqing Lu.
\newblock Learning feature embeddings for discriminant model based tracking,
  2020.

\bibitem{hierarchical}
Bineng Zhong, Bing Bai, Jun Li, Yulun Zhang, and Yun Fu.
\newblock Hierarchical tracking by reinforcement learning-based searching and
  coarse-to-fine verifying.
\newblock {\em {IEEE} Trans. Image Process.}, 28(5):2331--2341, 2019.

\bibitem{weakly}
Bineng Zhong, Hongxun Yao, Sheng Chen, Rongrong Ji, Tat{-}Jun Chin, and Hanzi
  Wang.
\newblock Visual tracking via weakly supervised learning from multiple
  imperfect oracles.
\newblock {\em Pattern Recognit.}, 47(3):1395--1410, 2014.

\bibitem{deepalignment}
Qinqin Zhou, Bineng Zhong, Yulun Zhang, Jun Li, and Yun Fu.
\newblock Deep alignment network based multi-person tracking with occlusion and
  motion reasoning.
\newblock {\em {IEEE} Trans. Multim.}, 21(5):1183--1194, 2019.

\bibitem{dasiamrpn}
Zheng Zhu, Qiang Wang, Bo Li, Wei Wu, Junjie Yan, and Weiming Hu.
\newblock Distractor-aware siamese networks for visual object tracking.
\newblock In {\em {ECCV} {(9)}}, volume 11213 of {\em Lecture Notes in Computer
  Science}, pages 103--119. Springer, 2018.

\end{thebibliography}
}

\end{document}